\documentclass{ieeeaccess}
\usepackage{cite}
\usepackage{amsmath,amssymb,amsfonts}
\usepackage{algorithmic}
\usepackage{graphicx}
\usepackage{textcomp}
\usepackage{url}

\newcommand{\etal}{\textit{et al.}}
\def\bs{\boldsymbol}

\makeatletter
\newcommand\footnoteref[1]{\protected@xdef\@thefnmark{\ref{#1}}\@footnotemark}
\makeatother

\def\BibTeX{{\rm B\kern-.05em{\sc i\kern-.025em b}\kern-.08em
    T\kern-.1667em\lower.7ex\hbox{E}\kern-.125emX}}
\begin{document}
\history{Date of publication xxxx 00, 0000, date of current version xxxx 00, 0000.}
\doi{10.1109/ACCESS.2017.DOI}

\title{Neural Image Compression and Explanation}
\author{
XIANG LI\authorrefmark{1}, AND SHIHAO JI\authorrefmark{1} (Member, IEEE)}

\address[1]{Department of Computer Science, Georgia State University, Atlanta, GA 30302, USA}

\markboth
{X. Li \headeretal: Neural Image Compression and Explanation}
{X. Li \headeretal: Neural Image Compression and Explanation}

\corresp{Corresponding author: Shihao Ji (sji@gsu.edu)}

\begin{abstract}
Explaining the prediction of deep neural networks (DNNs) and semantic image compression are two active research areas of deep learning with a numerous of applications in decision-critical systems, such as surveillance cameras, drones and self-driving cars, where interpretable decision is critical and storage/network bandwidth is limited. In this paper, we propose a novel end-to-end Neural Image Compression and Explanation (NICE) framework that learns to (1) explain the predictions of convolutional neural networks (CNNs), and (2) subsequently compress the input images for efficient storage or transmission. Specifically, NICE generates a sparse mask over an input image by attaching a stochastic binary gate to each pixel of the image, whose parameters are learned through the interaction with the CNN classifier to be explained. The generated mask is able to capture the saliency of each pixel measured by its influence to the final prediction of CNN; it can also be used to produce a mixed-resolution image, where important pixels maintain their original high resolution and insignificant background pixels are subsampled to a low resolution. The produced images achieve a high compression rate (e.g., about 0.6x of original image file size), while retaining a similar classification accuracy. Extensive experiments across multiple image classification benchmarks demonstrate the superior performance of NICE compared to the state-of-the-art methods in terms of explanation quality and semantic image compression rate. Our code is available at: \url{https://github.com/lxuniverse/NICE}.
\end{abstract}


\titlepgskip=-15pt

\maketitle

\section{Introduction}

Deep neural networks (DNNs) have become the de-facto performing technique in the field of computer vision~\cite{He_2016}, natural language processing~\cite{devlin2018bert}, and speech recognition~\cite{deepspeech3}. Given sufficient data and computation, they require only limited domain knowledge to reach state-of-the-art performance. However, the current DNNs are largely black-boxes with many layers of convolution, non-linearities, and gates, optimized solely for competitive performance, and our understanding of the reasoning of DNNs is rather limited. DNNs' predictions may be backed up by a claimed high accuracy on benchmarks. However, it is human's nature not to trust them unless human experts are able to verify, interpret, and understand the reasoning of the system. Therefore, the usage of DNNs in real world decision-critical applications, such as surveillance cameras, drones, autonomous driving, medicine and legal, still must overcome a trust barrier. To address this problem, researchers have developed many different approaches to explain the reasonings of DNNs~\cite{simonyan2013deep,zeiler2014visualizing,bach2015pixel,gan2015devnet,ribeiro2016should,dabkowski2017real,fong2017interpretable,li2016understanding,lundberg2017unified,shrikumar2017learning,YanRanRan18,HorUdwFla19}. Intuitively, interpretable explanations should be concise and coherent such that they are easier for human to comprehend. However, most existing approaches do not take these requirements into account as manifested by the opaqueness and redundancies in their explanations~\cite{simonyan2013deep,bach2015pixel,dabkowski2017real,fong2017interpretable}.

On the other hand, over 70\% of internet traffic today is the streaming of digital media, and this percentage keeps rising over years~\cite{cisco}. It has been challenging for classic compression algorithms, such as JPEG and PNG, to adapt to the growing demand. Recently, there is an increasing interest of using machine learning (ML) based approaches to improve the compression of images and videos~\cite{BalLapSim17,Prakash_2017,johnston2018improved,NakMaeMiy18}. Rather than using manually engineered basis functions for compression, these ML-based techniques learn semantic structures and basis functions directly from training images and achieve impressive performance compared to the classic compression algorithms. 

Usually, neural explanation and semantic image compression are addressed independently by two different groups of researchers. In light of the similarity between sparse explanation to image classification and sparse representation for image compression, in this paper we propose a deep learning based framework that integrates neural  explanation and semantic image compression into an end-to-end training pipeline. With this framework, we can train a sparse mask generator to generate a concise and coherent mask to explain the prediction of CNN; subsequently, this sparse mask can be used to generate a mixed-resolution image with a very high compression rate, superior to the existing semantic compression algorithms. This Neural Image Compression and Explanation (NICE) framework is critical to many real world decision-critical systems, such as surveillance cameras, drones and self-driving cars, that heavily rely on the deep learning techniques today. For these applications, the outputs of NICE: prediction, sparse mask / explanation, and the compressed mixed-resolution image can be stored or transmitted efficiently for decision making, decision interpretation and system diagnosis.

The main contributions of the paper are:
\begin{itemize}
	\item We propose a deep learning based framework that unifies neural explanation and semantic image compression into an end-to-end trainable pipeline, which produces prediction, sparse explanation and compressed images  at the same time;
	\item The proposed $L_0$-regularized sparse mask generator is trained in a weakly supervised manner without resorting to expensive dense pixel-wise annotations, and outperforms many existing explanation algorithms that heavily rely on  backpropagation;
	\item The proposed mixed-resolution image compression achieves a higher compression rate compared to the existing semantic compression algorithms, while retaining a similar classification accuracy with the original images.
	\item The proposed method is very efficient compared to the backpropagation-based alternatives, such as Saliency Map~\cite{simonyan2013deep} and CAM~\cite{CAM16}, as our method only requires forward propagation of the generator network. Experiments show that NICE is about 23x faster than Saliency Map, 16.5x faster than CAM and 2.8x faster than RTIS~\cite{dabkowski2017real}. This makes our method widely deployable in real-time applications.
\end{itemize}

\section{Related Work}
Our work is related to two active research areas of deep learning: neural explanation and semantic image compression. We therefore review them next.

\subsection{Neural Explanation}
In order to interpret DNN's prediction and gain insights of their operations, a variety of neural explanation methods have been proposed in recent years~\cite{simonyan2013deep,zeiler2014visualizing,bach2015pixel,gan2015devnet,ribeiro2016should,CAM16,dabkowski2017real,fong2017interpretable,li2016understanding,lundberg2017unified,selvaraju2017grad,shrikumar2017learning,YanRanRan18,HorUdwFla19,chen2018learning,bang2019explaining}. These methods can be categorized based on whether it is designed to explain the entire model behavior (global interpretability) or a single prediction (local interpretability)~\cite{xai19}. The goal of global interpretability is to identify predictor variables that best explain the overall performance of a trained model. This class of methods are crucial to inform population level decision for rule extraction or knowledge discovery~\cite{YanRanRan18,HorUdwFla19}. Local interpretability aims to produce interpretable explanations for each individual prediction and the interpretability occurs locally. Local interpretability is by far the most explored area of explainable AI~\cite{simonyan2013deep,ribeiro2016should,CAM16,lundberg2017unified,chen2018learning,bang2019explaining}. The primary idea is to measure a change of the final prediction with respect to changes of input or getting feature attribution for the final prediction. Different local explanation methods implement this idea in different ways. For example, occlusion-based explanation methods remove or alter a fraction of input data and evaluate its impact to the final prediction~\cite{dabkowski2017real,fong2017interpretable,li2016understanding,zeiler2014visualizing}. Gradient-based methods compute the gradient of an output with respect to an input sample by using backpropagation to locate salient features that are responsible to the prediction~\cite{bach2015pixel,gan2015devnet,selvaraju2017grad,shrikumar2017learning}. Other local interpretability methods explain data instances by approximating the decision boundary of a DNN with an inherently interpretable model around the predictions. For example, LIME~\cite{ribeiro2016should} and SHAP~\cite{lundberg2017unified} sample perturbed instances around a single data sample and fit a linear model to perform local explanations. RTIS~\cite{dabkowski2017real} extracts features from a DNN classifier and feeds extracted features and target label to an U-Net like generator to generate saliency maps for local explanations. L2X~\cite{chen2018learning} learns a stochastic map based on mutual information that selects instance-wise informative features. Built on top of L2X, VIBI~\cite{bang2019explaining} selects instance-wise key features that are maximally compressed about an input and informative about a decision based on an information-bottleneck principle. 

NICE falls in the category of local interpretability and aims to produce concise and coherent local explanations similar to Saliency Map~\cite{simonyan2013deep}, RTIS~\cite{dabkowski2017real} and VIBI~\cite{bang2019explaining}. But our method achieves briefness and comprehensiveness explicitly through an $L_0$-norm regularization and a smoothness constraint, optimized via stochastic binary optimization. 

The sparse mask generator of NICE is also related to a large body of research on semantic segmentation~\cite{long2015fully,he2017mask,chen2018encoder,papandreou2015weakly, pinheiro2015image, bearman2016s, Chen19ReDO}. In particular, our sparse mask generator is trained to maximize the final classification accuracy of the mixed-resolution images without resorting to expensive dense pixel-wise annotations. Therefore, it can be considered as a weakly supervised \textit{binary} segmenation algorithm that detects salient regions of an image. This is different to the existing semantic segmentation algorithms that employ different levels of supervision, such as full pixel-wise annotation~\cite{long2015fully,he2017mask}, image-level labels~\cite{papandreou2015weakly}, bounding boxes~\cite{dai2015boxsup}, scribbles~\cite{lin2016scribblesup}, points~\cite{bearman2016s}, or adversarial loss~\cite{Chen19ReDO}. Since the main goal of NICE is to provide a competitive or improved neural explanation, in our experiments we mainly compare NICE with deep explanation methods instead of segmentation algorithms. 

\subsection{Semantic Image Compression}
Classic image compression algorithms, such as JPEG~\cite{wallace1992jpeg} and PNG~\cite{sayood2002lossless}, have hard-coded procedures / components to compress images. For example, the JPEG compression first employs a discrete cosine transform (DCT) over each $ 8 \times 8$ image block, followed by quantization to represent the frequency coefficients as a sequence of binaries. The DCT can be seen as a generic feature extractor with a fixed set of basis functions that are irrespective of the distribution of the input images. Compared to standard image compression algorithms, the ML-based approaches~\cite{balle2016end,toderici2016,toderici2017full,theis2017lossy,BalLapSim17,johnston2018improved,li2018learning,NakMaeMiy18} can automatically discover semantic structures and learn basis functions from training images to achieve even higher compression rate. All of these ML-based approaches follow a similar structure of autoencoder, where an encoder is used to extract feature representation from images and a decoder is responsible to reconstruct images from the quantized representations. The main differences among these ML-based approaches are the architectures of encoder and decoder. While the majority of these algorithms~\cite{balle2016end,theis2017lossy,BalLapSim17,johnston2018improved,li2018learning,NakMaeMiy18} employ CNNs as the encoder and decoder, some others explore recurrent networks such as LSTM and GRU~\cite{toderici2016,toderici2017full}. 

To the best of our knowledge, all of these methods are not sufficiently content-aware, except the work~\cite{Prakash_2017} from Prakash \etal~which is probably the most relevant work to ours. While Prakash \etal~adopt CAM~\cite{CAM16} as the semantic region detector, we develop a principled $L_0$-regularized sparse mask generator to detect the semantic regions and further compress images with mixed resolutions. We will compare NICE with \cite{Prakash_2017} when we present results of semantic image compression.

\section{The NICE Framework}\label{sec:framework}
Given a training set $D = \{(\bs{x}_i, y_i), i = 1, 2,\cdots, N\}$, where $\bs{x}_i$ denotes the $i$-th input image and $y_i$ denotes the corresponding target, a neural network is a function $h(\bs{x}; \bs{\theta})$ parameterized by $\bs{\theta}$ that fits to the training data $D$ with the goal of achieving good generalization to unseen test data. To optimize $\bs{\theta}$, typically the following empirical risk minimization (ERM) is adopted:
\begin{equation}\label{eq:risk}
  \mathcal { R } ( \bs{\theta}) =  \frac { 1 } { N } \sum _ { i = 1 } ^ { N } \mathcal { L } \left( h \left( \bs { x } _ { i }; \bs { \theta } \right) , y_i \right),
\end{equation}
where $\mathcal{L}(\cdot)$ denotes the loss over training data $D$, such as the cross-entropy loss for classification or the mean squared error (MSE) for regression. The goal of this paper is to develop an approach that can explain the prediction of a neural network $h(\bs{x};\theta)$ in response to an input image $\bs{x}$; meanwhile, to reduce storage or network transmission cost of the image, we'd like to compress the image $\bs{x}$ based on the above derived explanation such that the compressed image $\tilde{\bs{x}}$ has the minimal file size while retaining a similar classification accuracy as the original image $\bs{x}$. 

\begin{figure}[thb]
	\begin{center}
		\includegraphics[width=1\linewidth]{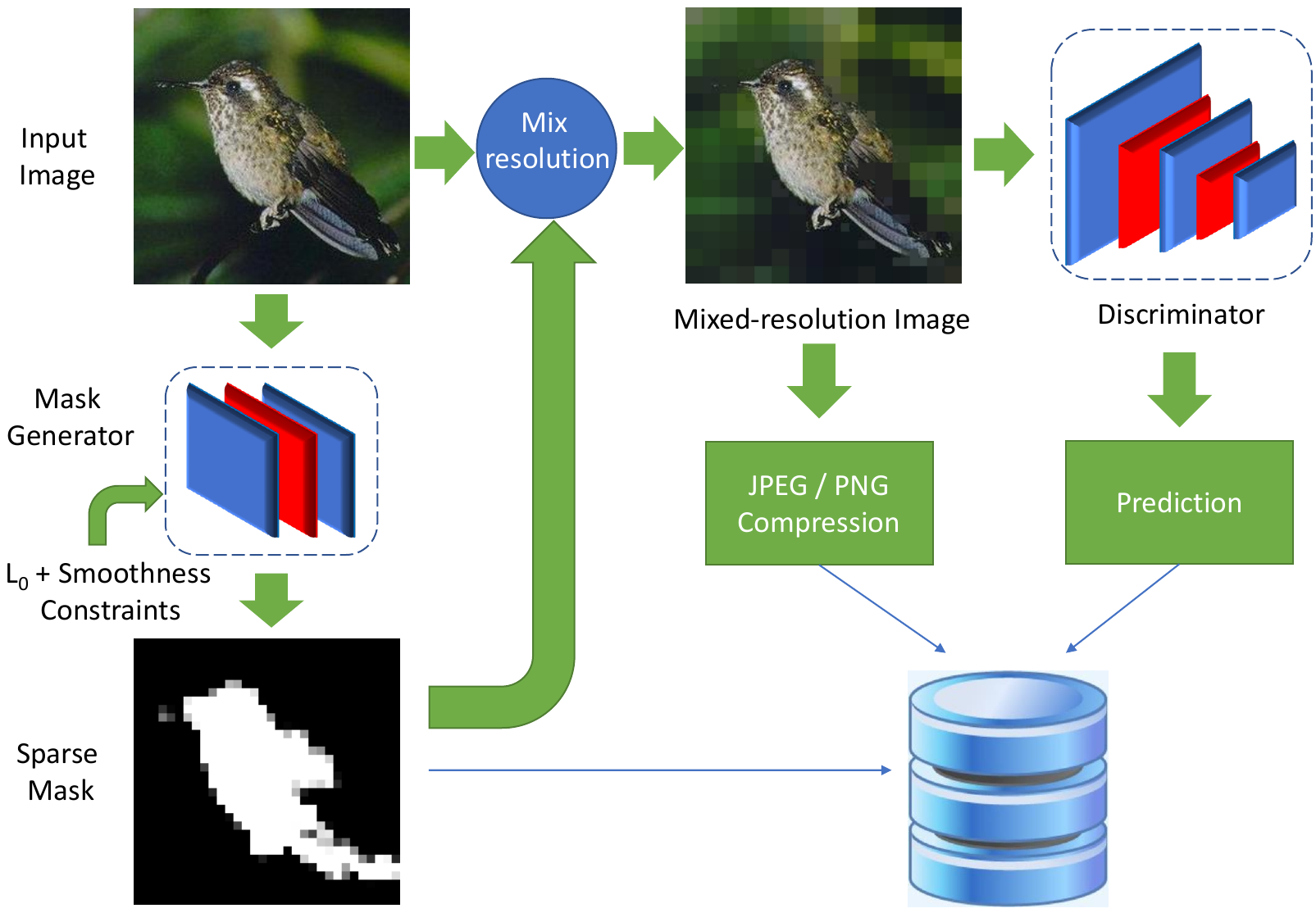}
	\end{center}\vspace{-0pt}
	\caption{Overall architecture of NICE.}
	\label{fig:structure}
\end{figure}

To meet these interdependent goals, we develop a Neural Image Compression and Explanation (NICE) framework that integrates explanation and compression into an end-to-end trainable pipeline as illustrated in Fig.~\ref{fig:structure}. In this framework, given an input image, a mask generator under the $L_0$-norm and smoothness constraints generates a sparse mask that indicates salient regions of the image. The generated mask is then used to transform the original input image to a mixed-resolution image that has a high resolution in the salient regions and a low resolution in the background. To evaluate the quality of sparse mask generator and the compressed image, at the end of the pipeline a discriminator network (e.g., CNN) classifies the generated image for prediction. Finally, the prediction, sparse mask and compressed image can be stored or transmitted efficiently for decision making, interpretation and system diagnosis. The whole pipeline is fully differentiable and can be trained end-to-end by backpropagation. We will introduce each of these components next.

\subsection{Sparse Neural Explanation}
To correctly classify an image, a state-of-the-art CNN classifier does not need to analyze all the pixels in an image. Partially, this is because not all the pixels in an image are equally important for image recognition. For example, although the background pixels may provide some useful clues to recognize an object, it is the pixels on the object that play a decisive role for recognition. Based on this understanding, we'd like to learn a set of random variables (one for each pixel of an image) such that the variables on object pixels receive high values while the variables on background pixels receive low values. In other words, we want to learn a binary segmentation model that can partition pixels into object pixels and background pixels. To make our segmentation discriminative,  we require the output of our model to be sparse/concise such that only the most important or influential pixels receive high values, and the remaining pixels receive low values. Furthermore, we expect the segmentation to be smooth/coherent within a small continuous region since most of natural objects usually have smooth appearances. We therefore request our neural explanation model to produce explanations that are concise and coherent. 
We will materialize these two requirements mathematically.

We model our neural explanation by attaching a binary random variable $z\in\{0,1\}$ to each pixel of an image:
\begin{equation}\label{eq:rationale}
  \tilde{\bs{x}}_i = \bs{x}_i \odot \bs{z}_i , \quad \bs{z}_i \in \{ 0,1 \}^P,
\end{equation}
where $\bs{z}_i$ denotes a binary mask for image $\bs{x}_i$, and $\odot$ is an element-wise product. Furthermore, we define $z_i^j$ the binary variable for pixel $j$ of image $\bs{x}_i$. We assume both image $\bs{x}_i$ and its mask $\bs{z}_i$ have the same spatial dimension of $m\times n$ or $P$ pixels. After training, we wish $z_i^j$ takes value 1 if pixel $j$ is on object and 0 otherwise.

We regard $\bs{z}_i$ as our explanation to the prediction of $h(\bs{x}_i;\bs{\theta})$ and learn $\bs{z}_i$ by minimizing the following $L_0$-norm regularized loss function:
\begin{align}\label{eq:loss_origin}
  \mathcal { R } ( \bs{\theta}, \bs{z} ) \!&= \! \frac { 1 } { N } \sum _ { i = 1 } ^ { N }\Big( \mathcal { L } \left( h \left( \bs { x } _ { i } \odot \bs { z } _ { i } ; \bs { \theta } \right) , y_i \right) \!  + \! \lambda ||\bs{z}_i||_0\Big) \\
  &=\!\frac { 1 } { N }\!\sum _ { i = 1 } ^ { N }\!\left(\! \mathcal { L } \left( h \left( \bs { x } _ { i } \odot \bs { z } _ { i } ; \bs { \theta } \right) , y_i \right) \!+\! \lambda \sum _ { j = 1 } ^ {    P   }  { 1 }_{\left[  z_i^j \neq 0 \right]}\!\right),\nonumber
\end{align}
where $\bs { 1 }_{[c]}$ is an indicator function that is $1$ if the condition $c$ is satisfied, and $0$ otherwise. Here, we insert~(\ref{eq:rationale}) into~(\ref{eq:risk}) and add an $L_0$-norm on the elements of $\bs{z_i}$, which \textit{explicitly} measures number of non-zeros in $\bs{z}_i$ or the sparsity of $\bs{z}_i$. By doing so, we'd like the masked image achieves the similar classification accuracy as the original image, while using as fewer pixels as possible. In other words, the sparse mask $\bs{z}_i$ can produce a concise explanation to the prediction of the classifier (i.e., the first requirement). To optimize~(\ref{eq:loss_origin}), however, we note that both the first term and the second term of~(\ref{eq:loss_origin}) are not differentiable w.r.t. $\bs { z }$. Therefore, further approximations need to be considered.

We can approximate this optimization problem via an inequality from stochastic variational optimization~\cite{SVO18}. Specifically, given any function $ \mathcal { F } (\bs{z})  $ and any distribution $q(\bs{z})$, the following inequality holds
\begin{equation}\label{eq:Ele}
        \min_{\bs{z}} \mathcal { F } (\bs{z}) \leq \mathbb{E}_{\bs{z}\sim q(\bs{z})} [\mathcal { F }(\bs{z})],
    \end{equation} 
i.e., the minimum of a function is upper bounded by the expectation of the function. With this result, we can derive an upper bound of~(\ref{eq:loss_origin}) as follows.

Since $z_i^j, \forall j\in\{1,\cdots,P\}$ is a binary random variable, we assume $z_i^j$ is subject to a Bernoulli distribution with parameter $\pi_i^j\in[0, 1]$, i.e. $z_i^j\sim \mathrm { Ber } (z;\pi_i^j)$. Thus, we can upper bound $\min_{\bs{z}} \mathcal { R } ( \bs { \theta }, \bs{z} )$ by the expectation
\begin{align}\label{eq:loss_1}
	\tilde{\mathcal { R }} ( \bs{\theta}, \bs{\pi} )\! & = \! \frac { 1 } { N }\! \sum _ { i = 1 } ^ { N } \Big(\mathbb { E } _ { q ( \bs {z}_i | \bs{\pi}_i ) } \Big[  \mathcal { L } \left( h \left( \bs { x } _ { i } \odot \bs { z } _ { i } ; \bs { \theta } \right) , y_i \right) \Big]  \nonumber \\ &  \quad + \lambda \sum _ { j = 1 } ^ { P } \pi_i^j\Big).
  \end{align}
Now the second term of~(\ref{eq:loss_1}) is differentiable w.r.t. the new model parameters $\bs{\pi}$. However, the first term is still problematic since the expectation over a large number of binary random variables $\bs{z}_i\in\{0, 1\}^P$ is intractable, so is its gradient. 

\subsubsection{The Hard Concrete Gradient Estimator}
Fortunately, this kind of binary latent variable models has been investigated extensively in the literature. There exist a numerous of gradient estimators to this problem, including REINFORCE~\cite{reinforce92}, Gumble-Softmax~\cite{gumbel-softmax17,concrete17}, REBAR~\cite{rebar17}, RELAX~\cite{relax18} and the hard concrete estimator~\cite{Louizos2017}, among which the hard concrete estimator is the one that is easy to implement and demonstrates superior performance in our experiments. We therefore resort to this gradient estimator to optimize~(\ref{eq:loss_1}). Specifically, the hard concrete gradient estimator employs a reparameterization trick to approximate the original optimization problem of~(\ref{eq:loss_1}) by a close surrogate loss function
\begin{align}\label{eq:loss_hc}
   \hat{\mathcal { R }} (\bs{\theta}, \!\log\bs{\alpha} ) \nonumber \!\! &=\!\! \frac { 1 } { N }\!\!\sum _ { i = 1 } ^ { N }\!\!\Bigg(\! \mathbb { E } _ {\bs{u}_i\sim \mathcal { U } ( 0,1 ) } \!\Big[\mathcal { L }\! \left( h \!\left( \bs { x } _ { i } \!\odot\! g(f(\log\bs{\alpha}_i,\!\bs{u}_i)) ; \bs { \!\theta } \right) \!,\! y_i \right) \!\!\Big] \!  \nonumber\\ & \quad   +  \lambda\sum _ { j = 1 } ^ { P }\! \sigma\!\left(\!\log \alpha_i^j - \beta \log \frac { - \gamma } { \zeta } \right)\!\!\!\Bigg) \nonumber\\
  	& = \mathcal{L}_{D}(\bs{\theta}, \log\bs{\alpha}) + \lambda \mathcal{L}_{C}(\log\bs{\alpha}),
\end{align}
with 
\vspace{-10pt}
\begin{align}\label{eq:hard_concrete}
  f(\log\bs{\alpha}_i, \bs{u}_i) & = \sigma \left( ( \log\bs{u}_i - \log ( 1 - \bs{u}_i ) \right.\nonumber \\ &  \quad \left. + \log\bs{\alpha}_i ) / \beta \right) ( \zeta - \gamma) + \gamma, 
\end{align}
and
\begin{align}\label{eq:hard_concrete2}
  g(\cdot)= \min ( 1 , \max ( 0 , \cdot ) ),
\end{align}
where $\sigma(t)=1/(1+\exp(-t))$ is the sigmoid function, $\mathcal{L}_D$ measures how well the classifier fits to training data $D$, $\mathcal{L}_C$ measures the expected number of non-zeros in $\bs{z}$, and $\beta=2/3$, $\gamma = -0.1$ and $\zeta = 1.1$ are the typical parameters of the hard concrete distribution. Function $g(\cdot)$ is a hard-sigmoid function that bounds the stretched concrete distribution between 0 and 1. For more details on the hard concrete gradient estimator, we refer the readers to~\cite{Louizos2017}. With this reparameterization, the surrogate loss function~(\ref{eq:loss_hc}) is differentiable w.r.t. its parameters.

\subsubsection{Smoothness Regularization}
The $L_0$-regularized objective function developed above enforces the sparsity/conciseness of an explanation. To improve the coherence of an explanation, we introduce an additional smoothness constraint on the mask:
\begin{align}\label{eq:loss3}
	\mathcal { L } _ {S}(\log\bs{\alpha}) 
	 &= \frac { 1 } { N }  \sum _ { i = 1 } ^ { N }\mathbb { E } _ { q ( \bs{z}_i|\log\bs{\alpha}_i ) } \bigg[ \sum _ { m, n = 1 } ^ { w, h } \!\!\!\Big( \left| z_i^{m, n}\!\!\! - z_i^{m-1, n} \right| \nonumber \\ 
	 & \!\!\!\!\!\!\!\!\!\!\!\!\!\!\!\!\!\!\!\! +\!\! \left| z_i^{m, n}\!\!\! -\!\! z_i^{m, n-1} \right|\!\! +\!\! \left| z_i^{m, n}\!\!\! - \!\! z_i^{m-1, n-1} \right| \nonumber  
	 \!\!+\!\! \left| z_i^{m, n}\!\!\! - z_i^{m-1, n+1} \right| \Big) \bigg] \nonumber\\ 
	 &\!\!\!\!\!\!\!\!\!\! \approx\!  \frac { 1 } { N }\!\sum _ { i = 1 } ^ { N }\sum _ { m, n = 1 } ^ { w, h }\!\! \left( \left| y_i^{m, n}\!\! - y_i^{m-1, n} \right| \!+\! \left| y_i^{m, n}\!\! - y_i^{m, n-1} \right| \! \right. \nonumber\\ 
	 & \!\!\!\!\!\! \left. + \! \left| y_i^{m, n}\!\! - y_i^{m-1, n-1} \right| \!+\! \left| y_i^{m, n}\!\! - y_i^{m-1, n+1} \right| \right),
\end{align}
where $y_i^{m, n}$ is the expectation of random variable $z_i^{m, n}$ under the hard concrete distribution $q(\bs{z}_i|\log\bs{\alpha}_i)$, which can be calculated as:
\begin{equation}\label{eq:exp_concrete}
  y = \mathbb{E}_{q(z|\log\alpha)}[z]=\sigma\left( \log \alpha - \beta \log \frac { - \gamma } { \zeta } \right).
\end{equation}
Note that this smoothness constraint penalizes the discrepancy of $z$ among its four neighborhoods, and thus a coherence explanation is preferred (i.e., the second requirement). To avoid notational clutter, in~(\ref{eq:loss3}) some of the boundary conditions are not rigorously checked, but we hope they will be apparent given the context. With this additional regularization, our final objective is then a composition of three terms
\begin{equation}\label{eq:overall_loss}
  \mathcal { L }(\bs{\theta}_d, \log\bs{\alpha})  = \mathcal { L } _ {D} + \lambda_1 \mathcal { L } _ { C } +\lambda _ {2} \mathcal { L } _ { S },
\end{equation}
where $\lambda_1$ and $\lambda _ {2}$ are the regularization hyperparameters that balance the data loss $\mathcal{L}_D$, the capacity loss $\mathcal{L}_C$ and the smoothness loss $\mathcal{L}_S$. It is worthy noting that from now on we denote the parameters of classifier (discriminator) $\bs{\theta}_d$ to distinguish it from the parameters of generator $\bs{\theta}_g$ that will be introduced next. 

After training, we get $\log\bs{\alpha}$ for each input image $\bs{x}$. At testing time, we employ the following estimator to generate a sparse mask:
\begin{equation}\label{eq:testing}
  \hat{\bs{z}} = \min(\bs{1} , \max(\bs{0}, \sigma\left((\log\bs{\alpha})/\beta\right)(\zeta - \gamma) + \gamma)),
\end{equation}
which is the sample mean of $\bs{z}$ under the hard concrete distribution $q(\bs{z}|\log\bs{\alpha})$.

\subsection{Semantic Image Compression}
Upon receiving the sparse mask $\hat{\bs{z}}$ from above, we can use it to generate a mixed-resolution image for semantic image compression, as shown in Fig.~\ref{fig:structure}. Suppose that we have an input image $\bs{x}$ and a sparse mask $\hat{\bs{z}}\in[0, 1]^P$, a mixed-resolution image can be generated by
\begin{equation}\label{eq:mix_resolution}
  \tilde{\bs{x}} = M(\bs{x},\hat{\bs{z}}) = \bs{x} \odot \hat{\bs{z}} + \bs{x} _ {b} \odot (1 - \hat{\bs{z}}),
\end{equation}
where $\bs{x}_b$ is a low resolution image that can be generated by subsampling original image $\bs{x}$ with a block size of $b\times b$, which can be efficiently implemented by average pooling with a $b\times b$ filter and a stride of $b$. Here $b$ is a tunable hyperparameter that trades off  between the image compression rate and the classification accuracy of the classifier. In other words, the larger $b$ is, the lower resolution images will be generated and thus a lower classification accuracy, and vice-versa. As we can see, when $b=1$ the mixed-resolution image $\tilde{\bs{x}}$ is equal to the original image $\bs{x}$; when we use the image size as $b$, the mixed-resolution image $\tilde{\bs{x}}$ becomes a masked image with a constant value as background. When $b$ is a value between these two extremes, we can generate mixed-resolution images of different levels of quality.

\subsection{Sparse Mask Generator}
The learning of sparse mask $\bs{z}$ discussed above is transductive, by which we can learn a mask for each image in training set $D$. However, this approach cannot generate masks for new images that are not in the training set $D$. A more desirable approach is inductive, which can be implemented through a generator $G(\bs{x};\bs{\theta}_g)$ such that it can produce a sparse mask given any image $\bs{x}$ as input. We model this generator as a neural network parameterized by $\bs{\theta}_g$. 

To integrate this generator into an end-to-end training pipeline, we model this generator to output $\log\bs{\alpha}$ given an input image $\bs{x}$; we can then sample a sparse mask $\bs{z}$ from the hard concrete distribution $q(\bs{z}|\log\bs{\alpha})$, i.e., $\bs{x}\xrightarrow{G(\cdot;\bs{\theta}_g)}\log\bs{\alpha}\xrightarrow{\text{sample}}\bs{z}$. With this reparameterization, the overall loss function~(\ref{eq:overall_loss}) becomes $\mathcal{L}(\bs{\theta}_d, \bs{\theta}_g)$, which can be minimized by optimizing the generator network $\bs{\theta}_g$ and the discriminator network $\bs{\theta}_d$ jointly with backpropagation. In the experiments, we employ a CNN as our sparse mask generator as CNN is the de-facto technique today for image related analysis.

\section{Experiments}
To evaluate the performance of NICE, we conduct extensive experiments on three image classification benchmarks: MNIST~\cite{lecun1998gradient}, CIFAR10~\cite{cifar10} and Caltech256~\cite{griffin2007caltech}~\footnote{\url{http://www.vision.caltech.edu/Image_Datasets/Caltech256/}}. Since NICE is a neural explanation and semantic compression algorithm, we compare NICE with the state-of-the-art algorithms in neural explanation and semantic compression. For neural explanation, we compare NICE with Saliency Map~\cite{simonyan2013deep}, RTIS~\cite{dabkowski2017real} and CAM~\cite{CAM16} via visualization and the post-hoc classification. For semantic image compression, we compare NICE with the CAM-based method proposed in~\cite{Prakash_2017}, a state-of-the-art semantic compression algorithm that is the most relevant to ours. 



\subsection{Implementation Details} \label{sec:imp}

\subsubsection{Image Classification Benchmarks}
\label{appendix:exp}
MNIST~\cite{lecun1998gradient} is a gray-level image dataset containing 60,000 training images and 10,000 test images of the size $28 \times 28$ for handwritten digits classification. CIFAR10~\cite{cifar10} contains 10 classes of RGB images of the size $ 32\times32$, in which 50,000 images are for training and 10,000 images are for test. Caltech256~\cite{griffin2007caltech} is a high-resolution RGB image dataset containing 22,100 images from 256 classes of man-made and natural objects, such as plants, animals and buildings, etc. Since MNIST and CIFAR10 are low-resolution images, we use them mainly to demonstrate NICE's performance on neural explanation. For the high-resolution images of Caltech256, we demonstrate NICE's performance on neural explanation and semantic image compression. 

\subsubsection{Network Architectures and Training Details}
\label{appendix:arch}
The network architectures of the sparse mask generators and CNN classifiers (discriminators) used in the MNIST, CIFAR10 and  Caltech256 experiments are provided in Table~\ref{tab:arch}.

\begin{table}[th!]
	\centering
	\caption{Network architectures of the generators and discriminators used in the experiments. Layer abbreviations used in the table: [C: Convolution; R: Relu; M: MaxPooling; Up: UpSample].} 	\vspace{-5pt}
    \begin{tabular}{|l|l|l|}
    \hline
    Dataset & Generator & Discriminator \\ \hline
    MNIST & C(1,1,3,1,1) & LeNet5-Caffe  \\ \hline
    
    CIFAR10 & C(1,1,5,1,0)  + M(2) + Up(2)  & VGG11 + FC(512, 10)  \\ \hline
    
    Caltech256 & C(3,1,3,1,1) + R + M(2) & ResNet18  \\
    & C(1,1,3,1,1) + R + M(2) & FC(512, 256)       \\
    & C(1,1,3,1,1) + M(2) + Up(8)    &    \\ \hline
	\end{tabular}
	\vspace{-0pt}
    \label{tab:arch}\vspace{-5pt}
\end{table}

We pretrain three CNN classifiers (discriminators) on the three image classification benchmarks: MNIST, CIFAR10 and Caltech256 and achieve the classification accuracies of 99\%, 90.8\% and 78.3\%, respectively. These classifiers are the target CNNs we aim to explain. The architectures of the generators are tuned by us through extensive architecture search.  The hyperparameters $\lambda_1$ and $\lambda_2$ in the overall loss (\ref{eq:overall_loss}) are tuned on validation set to balance the classification accuracy and sparsity/smoothness of the masks.

In the MNIST experiments, different $\lambda_1$s are used to generate sparse masks with different percentages of non-zeros (sparse explanations). $\lambda_2$ is set to 0 for all the MNIST experiments as the algorithm can generate coherent explanations without the smoothness constraint. The block size of the low resolution image $\bs{x}_{b}$ is set to 28, which means a constant background is used to generate the mixed-resolution images. We use the Adam optimizer~\cite{kingma2014adam} with a learning rate of 0.001 and a decay rate of 0.1 at every 5 epochs.

In the CIFAR10 experiments, the block size of the low resolution image $\bs{x}_b$ is set to 32, thus a constant background image is used to generate the mixed-resolution images. We set $\lambda_1=3$ and $\lambda_2=0.01$ and train the pipeline by using the Adam optimizer with a learning rate of 0.001 and a decay rate 0.1 at every 5 epochs.

In the Caltech256 experiments, we split the dataset into a training set of 16,980 images and a test set of 5,120 images\footnote{20 images per class are included in the test set.}, where 5,120 images in training set is first used as validation set for architecture search and hyperparameter tuning and later the full 16,980 training images are used to train the final pipeline. The images are resized to $256 \times 256$ as inputs. We set $b=256$ to generate the lowest resolution images $\bs{x}_{b}$, and set $\lambda_1=5$ and  $\lambda_2=0.01$ and train the pipeline by using the SGD optimizer with a learning rate of 0.001 and a cosine decay function. 

On different datasets, we experiment with different optimizers. The best performing one is selected based on its performance on validation set. It turns out that SGD works better on Caltech256, while Adam works better on MNIST and CIFAR10. 

\subsection{Explaining CNN's Predictions}
We first demonstrate NICE on explaining the predictions of the target CNNs we pretrained above. To do so, we incorporate the target CNN as discriminator into the pipeline (Fig.~\ref{fig:structure}), and freeze its parameters $\bs{\theta}_d$ and only update the parameters of generator $\bs{\theta}_g$ by optimizing the overall loss~(\ref{eq:overall_loss}). The sparse mask $\bs{z}$ generated by the generator serves as the explanation to CNN's prediction since the mask indicates the salient region that has strong influence to the final prediction.

\subsubsection{MNIST}

\begin{figure}[t]
	\begin{center}
		\includegraphics[width=1\linewidth]{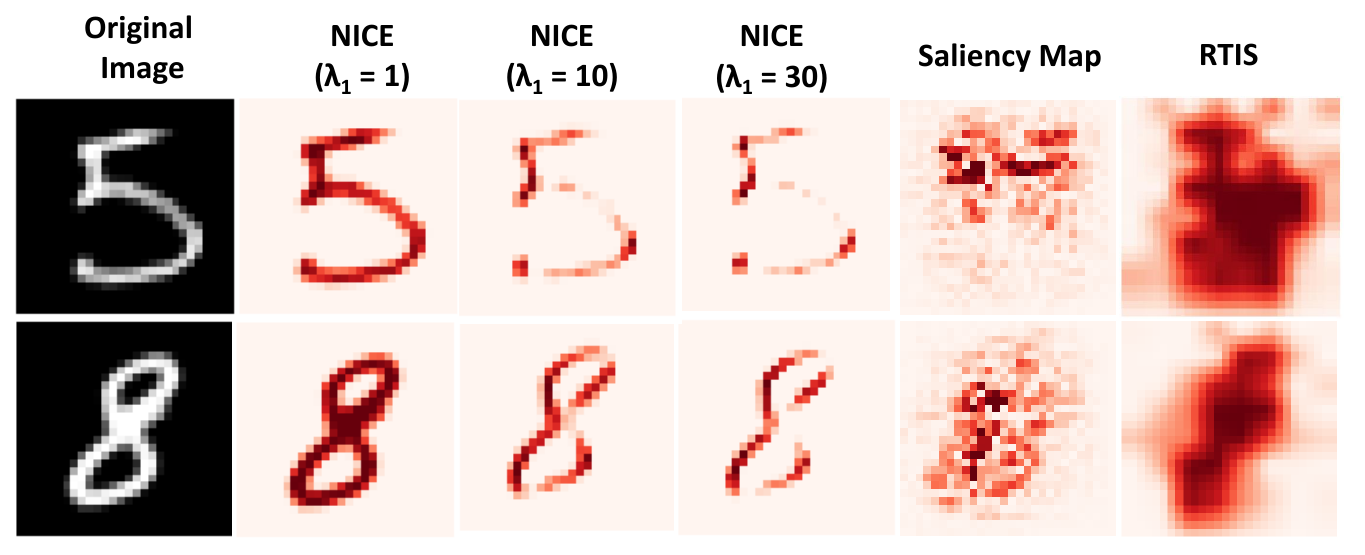}
	\end{center}\vspace{-5pt}
  \caption{The sparse masks generated by NICE, Saliency Map~\cite{simonyan2013deep} and RTIS~\cite{dabkowski2017real} on the MNIST dataset. The dark red color represents high values (close to 1), indicating strong influence to the final decisions. By adjusting $\lambda_1$ of NICE, we can control the sparsity of the explanations.}\vspace{-10pt}
	\label{fig:mnist_exp}
\end{figure}

We train the NICE pipeline on the MINST dataset to explain the prediction of the target LeNet5 classifier we pretrained above. Fig.~\ref{fig:mnist_exp} illustrates example sparse explanations generated by NICE with different $\lambda_1$s (when $\lambda_2=0$). As we can see, when $\lambda_1$ increases, the amount of non-zeros in the mask $\bs{z}$ decreases and NICE can produce sparser explanations to the final predictions. When $\lambda_1=1$, the explanations are almost identical to the input images, and when $\lambda_1=30$, the masks identify sparser but more influential regions for the final predictions. As a comparison, we also include the explanation results produced by Saliency Map~\cite{simonyan2013deep} and RTIS~\cite{dabkowski2017real}~\footnote{\label{cam_note}CAM~\cite{CAM16} does not perform well on small images, and we observe no published work provides CAM's results on MNIST and CIFAR10. We therefore ignore its results here as well.}. While NICE highlights coherent regions over digits as explanations, Saliency Map, a backpropagation-based approach, identifies discontinued regions as explanations, which are quite blurry and difficult to understand. RTIS can yield coherent regions as explanations but the regions identified are overly smooth. Apparently, the explanations produced by NICE are more concise, coherent and match well with how humans explain their own predictions. 

\subsubsection{CIFAR10}

\begin{figure*}[th!]
	\begin{center}
		\includegraphics[width=1\linewidth]{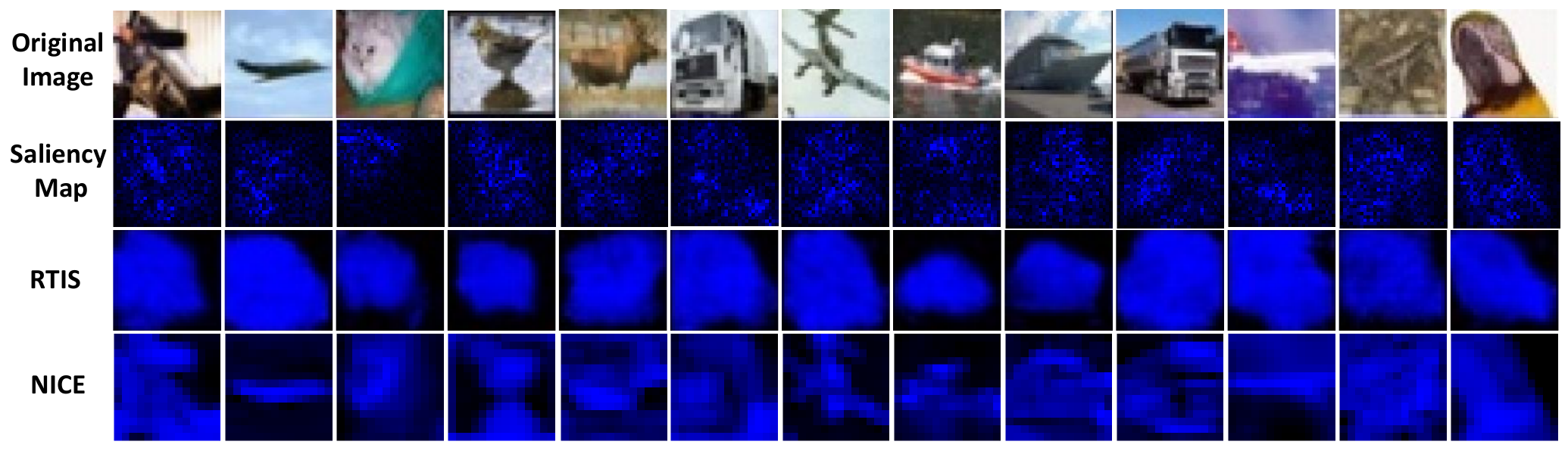}
	\end{center}\vspace{-5pt}
	\vspace{-5pt}
	\caption{Comparison of explanations generated by Saliency Map~\cite{simonyan2013deep}, RTIS~\cite{dabkowski2017real} and NICE on some CIFAR10 images. The RTIS results are from the RTIS paper. Compared to Saliency Map and RTIS, the explanations generated by NICE are more concise and the boundaries of salient regions are much sharper.}
	\label{fig:cifar_exp}\vspace{-5pt}
\end{figure*}

We also train the NICE pipeline on the CIFAR10 dataset to explain the target VGG11 classifier we pretrained above. Fig.~\ref{fig:cifar_exp} compares the explanations produced by Saliency Map~\cite{simonyan2013deep}, RTIS~\cite{dabkowski2017real} and NICE on some CIFAR10 images~\footnoteref{cam_note}. The RTIS results are directly cited from the RTIS paper, and we apply Saliency Map and  NICE on the same set of CIFAR10 images selected by the RTIS paper. Due to the low resolution of the images, it's very challenging to generate reliable explanations. As we can see, the explanations generated by NICE are more concise and the boundaries of salient regions are much sharper than those of Saliency Map and RTIS. The superior performance of NICE is most likely due to the $L_0$-norm regularization that \emph{explicitly} promotes the sparsity of an explanation.

\subsubsection{Caltech256}

\begin{figure}[h!]
	\begin{center}
		\includegraphics[width=1\linewidth]{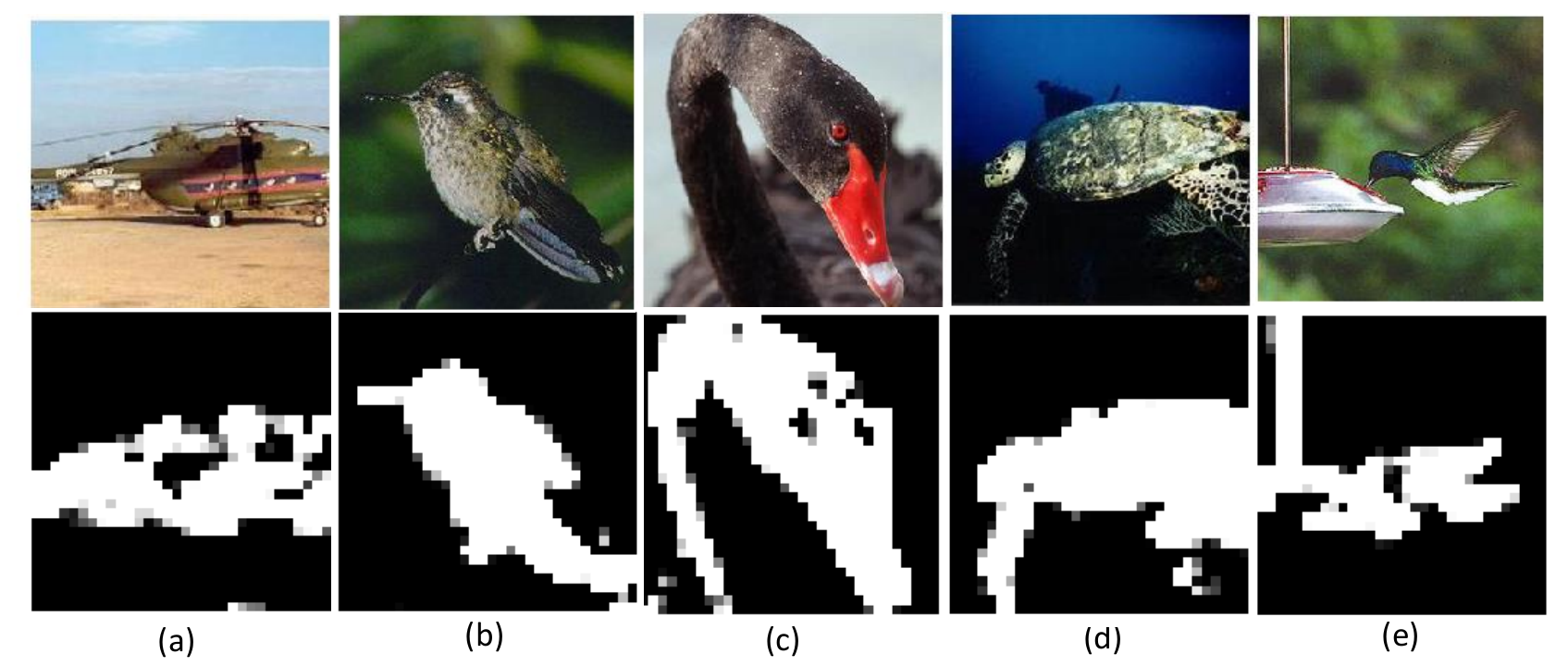}
	\end{center}\vspace{-5pt}
	\caption{The sparse masks generated by NICE on Caltech256 images. The predictions are correct to (a,b,c,d) and incorrect to (e). Even though the prediction is incorrect, the sparse mask (e) provides an intuitive explanation why the discriminator predicts an image of ``humming bird" as ``bread maker".}\vspace{-5pt}
	\label{fig:caltech_exp}
\end{figure}

Similarly, we train the NICE pipeline on the Caltech256 dataset to explain the predictions of the ResNet18 classifier we pretrained above. Fig.~\ref{fig:caltech_exp} demonstrates the sparse masks produced by NICE for different images in Caltech256. As we can see, the generated explanations are very concise and coherent, i.e., the sparse masks are mainly concentrated on the object regions, which align very well with our reasoning on these images. Additionally, the generated sparse masks also provide intuitive explanations when the classifier makes mistakes. For example, as shown in Fig.~\ref{fig:caltech_exp}(e), the classifier incorrectly predicts an image of ``humming bird" as ``bread maker". The corresponding sparse explanation highlights the influential regions contributing the most to the classifier's prediction. Clearly, the classifier utilizes both the regions of the humming bird and the bird-feeder for the prediction, and the combination of the two regions confuses the classifier and leads to the incorrect classification. Such an explanation is very useful for system diagnosis: it uncovers the vulnerabilities and flaws of the classifier, and can help to improve the performance of the system.

\begin{figure}[h!]
	\begin{center}
		\includegraphics[width=1\linewidth]{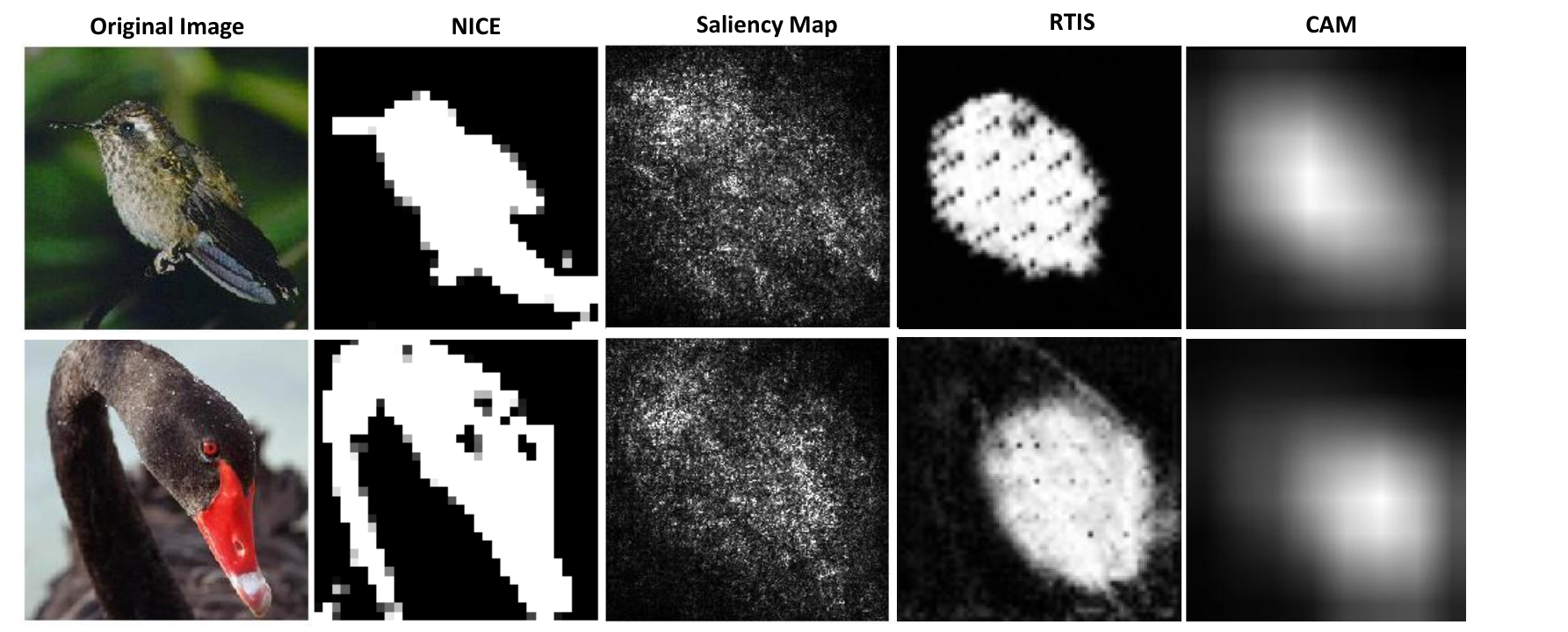}
	\end{center}\vspace{-5pt}
	\caption{The sparse masks generated by NICE, Saliency Map~\cite{simonyan2013deep}, RTIS~\cite{dabkowski2017real} and CAM~\cite{CAM16} on the Caltech256 dataset. NICE highlights the whole body of object as the explanation instead of edges or scattered pixels as identified by Saliency Map, or overly-smooth regions as identified by RTIS and CAM.}
	\vspace{-5pt}
	\label{fig:explains}
\end{figure}

Fig.~\ref{fig:explains} illustrates the comparison of NICE with Saliency Map~\cite{simonyan2013deep}, RTIS~\cite{dabkowski2017real} and CAM~\cite{CAM16} on the Caltech256 images. As we can see, our algorithm highlights the whole body of object as the explanation while Saliency Map typically identify edges or scattered pixels as the explanation. RTIS and CAM can identify coherent salient regions of an image, however, those regions are overly smooth and cover large background regions. Moreover, the saliency maps generated by RTIS usually have some black grids in the highlighted parts, which are caused by the upsampling step in the mask generator. Apparently, our explanations are more concise and coherent than those of the competing methods, and can preserve semantic contents of the images with a high accuracy. The superior performance of NICE on identifying semantic regions plays a critical role in semantic image compression as we will demonstrate later.

\begin{figure}[h!]
	\begin{center}
		\includegraphics[width=1\linewidth]{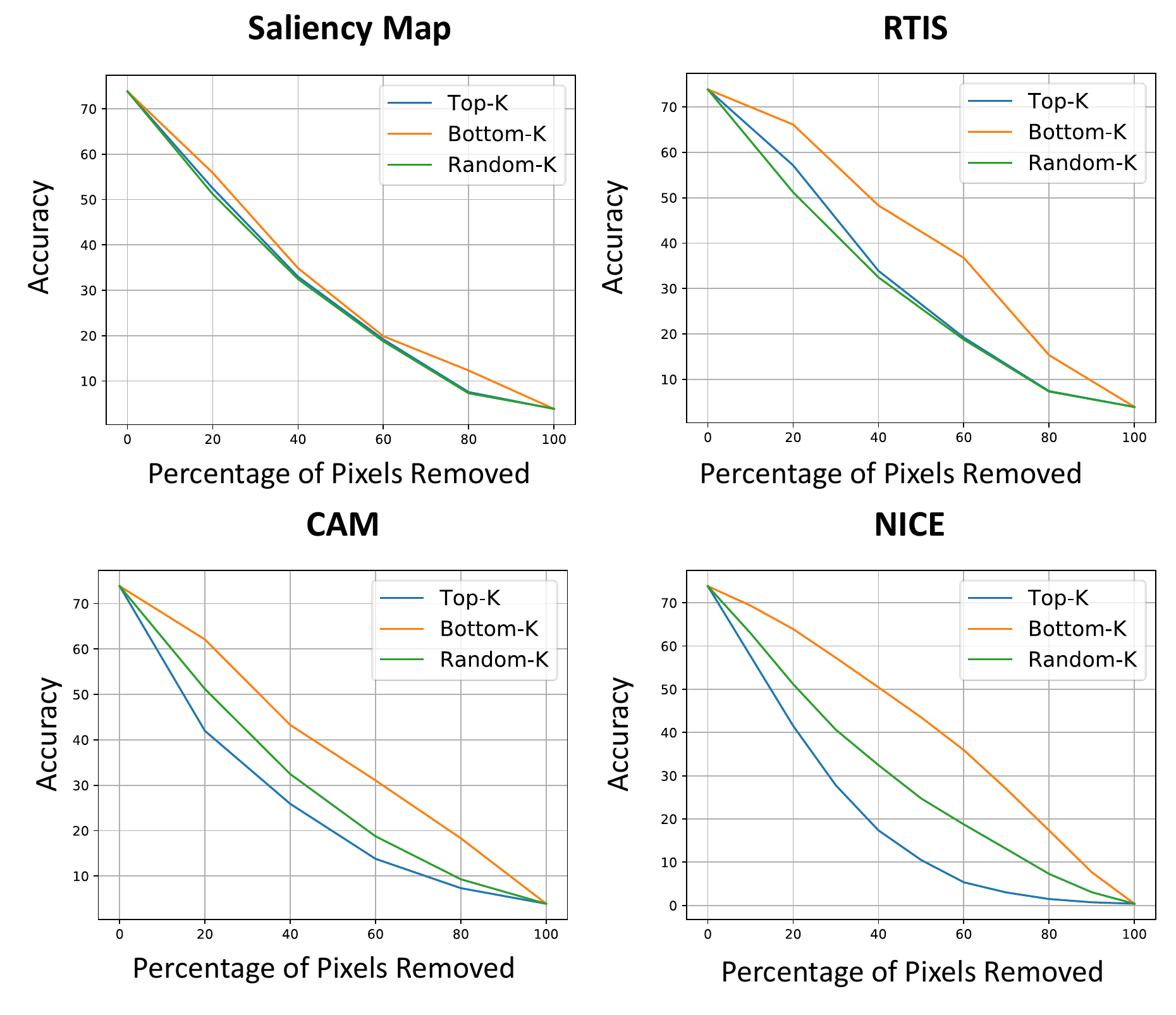}
	\end{center}\vspace{-5pt}
	\caption{The evolution of classification accuracies on the Caltech256 test dataset when different percentages of pixels are filled with random values.} 
	\label{fig:curves_remove}
\end{figure}

To evaluate NICE’s performance of identifying important pixels from an image, Fig.~\ref{fig:curves_remove} demonstrates the evolution of classification accuracies on the Caltech256 test dataset when different percentages of pixels are filled with random values sampled uniformly from $[0,255]$ (a.k.a. post-hoc classification evaluation). We compare three different strategies of selecting pixels for random value imputation: (1) Top-K\% pixels sorted descending by $\log\alpha_j, \forall j\in\{1,2,\cdots,P\}$, (2) Bottom-K\% pixels sorted descending by $\log\alpha_j$, and (3) uniformly random K\% pixels. Similarly, the same post-hoc classification evaluation is performed with Saliency Map, RTIS and CAM. As we can see from Fig.~\ref{fig:curves_remove}, NICE identifies important pixels from images as randomizing their Top-K\% values incurs a dramatic accuracy loss compared to random pixel selection or Bottom-K\% pixel selection. The results of Saliency Map, RTIS and CAM show insignificant accuracy loss when randomizing their Top-K\% pixels, demonstrating the superior performance of NICE on identifying salient regions.

\subsubsection{Transferability of Sparse Mask Generator}
\begin{figure}[t]
	\begin{center}
		\includegraphics[width=1\linewidth]{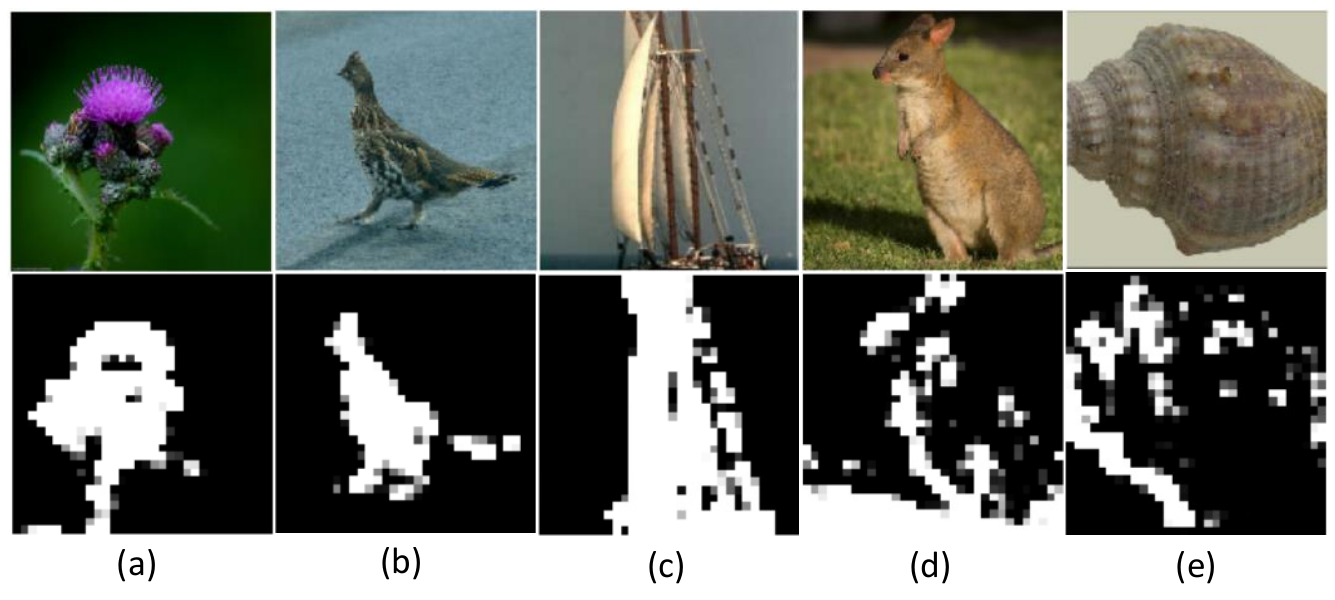}
	\end{center}\vspace{-5pt}
	\caption{Sample ImageNet images and their sparse masks generated by the generator trained on Caltech256. While the ground truth labels of (a, b, c) are included in Caltech256, the ground truth labels of (d, e) are not in Caltech256. NICE is able to generate accurate sparse masks for images in (a, b, c). But when the classes are not in Caltach256 the masks are not very accurate as shown in (d, e).}\vspace{-5pt}
	\label{fig:imagenet}
\end{figure}

The experiments above demonstrate the superiority of the sparse mask generator in generating concise and coherent explanations to target classifier's predictions. This has been verified in the case when the generator is applied to the test images from the same dataset. Since our generator is inductive, it would be interesting to test if a generator trained from one dataset could be applied to images from other datasets, which have similar statistics but yet have some mismatch, i.e., the transferability of generator.

To measure the transferability of generator, we apply the generator trained on Caltech256 to the ImageNet images~\cite{deng2009imagenet}. Although both Caltech256 and ImageNet contain high-resolution RGB images, ImageNet contains 1000 classes which is 4 times of Caltech256's and the ImageNet images tend to be more complex than those in Caltech256. To feed the ImageNet images to the generator trained on Caltech256, we resize the ImageNet images to $256\times 256$. Fig.~\ref{fig:imagenet} illustrates some sample ImageNet images and their sparse masks. Images (a, b, c) are from the classes that included in Caltech256, while images (d, e) are from the classes that are not in Caltech256. It shows that for the classes that overlap with Caltech256, the generator can generate sparse masks that align well with the object in the images, while for the images that are from non-overlapping classes the masks are not very accurate, indicating the transferability of generator is class dependent.

%

\subsection{Semantic Image compression }\label{sec:image_compression}

\begin{figure}[h!]
	\begin{center}
		\includegraphics[width=0.95\linewidth]{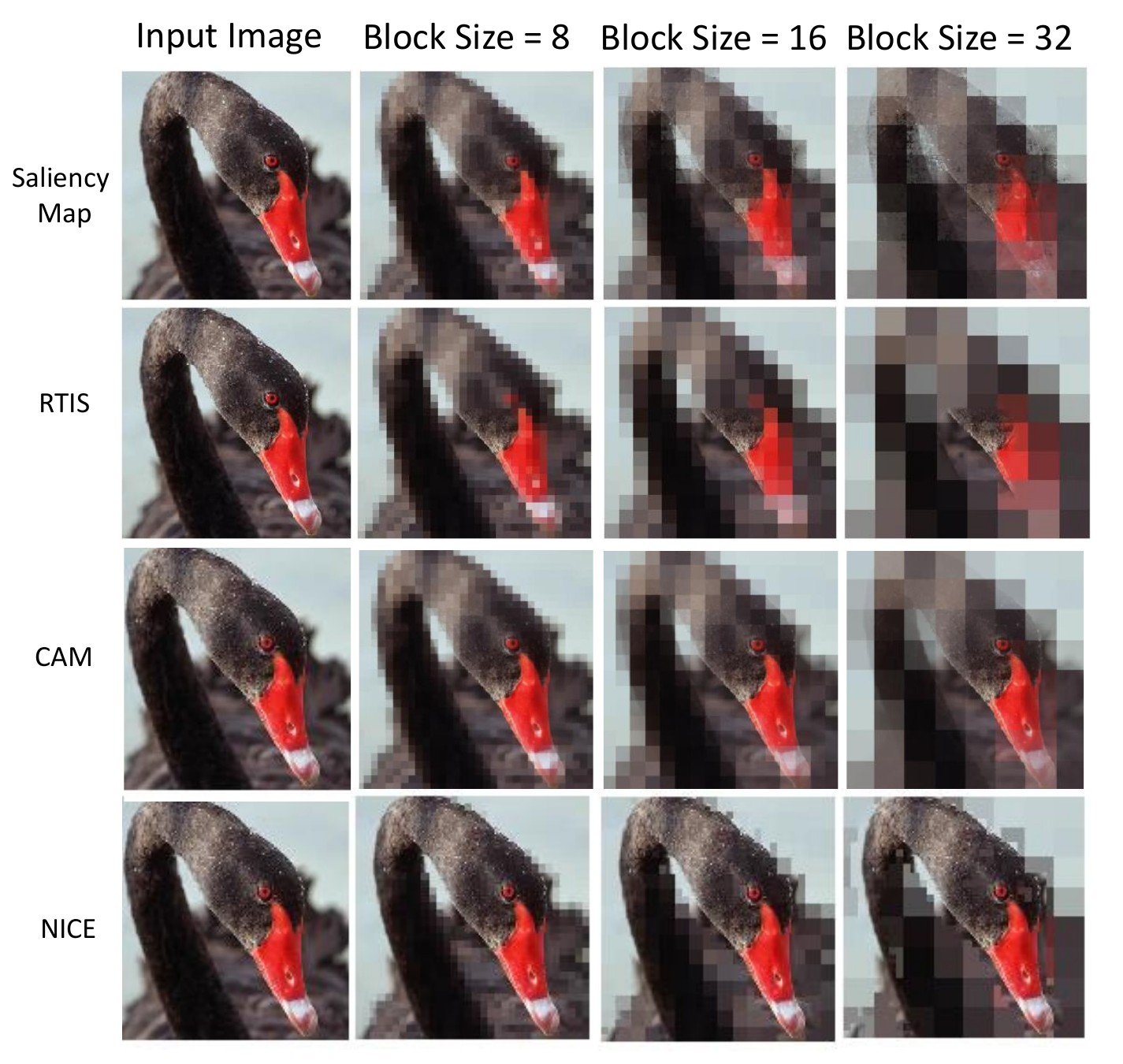}
	\end{center}\vspace{-5pt}
	\caption{The mixed-resolution images generated by NICE, Saliency Map, RTIS and CAM with different block size $b$s.}\vspace{-5pt}
	\label{fig:mix}
\end{figure}


Finally, we evaluate the semantic image compression performance of NICE on the Caltech256 images. As a comparison, we also use the salient regions generated by Saliency Map, RTIS and CAM for semantic image compression. In this task, two approaches can be used to train the NICE pipeline: (1) \textbf{Discriminator-fixed}: given a pretrained discriminator, we freeze its parameters $\bs{\theta}_d$ in the pipeline and only update the parameters of generator $\bs{\theta}_g$ by optimizing the overall loss~(\ref{eq:overall_loss}). In this case, the mask generator is trained to generate sparse explanation to the original discriminator. (2) \textbf{Discriminator-finetuned}: similar to discriminator-fixed except that the top few layers of the discriminator $\bs{\theta}_d$ are finetuned. In this case, the discriminator can adjust its parameters to improve its predictions on the mixed-resolution images, and thus higher accuracy and compression rate are expected. Note that due to their specific training methodologies, Saliency Map, RTIS and CAM do not have the flexibility of finetuning their discriminators, limiting their applications to semantic compression tasks.

As a start, we use the sparse masks generated by NICE, Saliency Map, RTIS and CAM to produce a set of mixed-resolution images via~(\ref{eq:mix_resolution}) for visualization. Fig.~\ref{fig:mix} illustrates some example mixed-resolution images generated with different algorithms and block size $b$s. As we can see, NICE generated mixed-resolution images are clearly better than those from other algorithms. Thanks to the high accuracy of NICE on identifying salient regions, even when the background regions are subsampled with a block size of 32, the discriminator can still successfully classify these images. As a result, high compression rate and high classification accuracy can be achieved simultaneously. 

\begin{figure}[t]
	\begin{center}
		\includegraphics[width=1\linewidth]{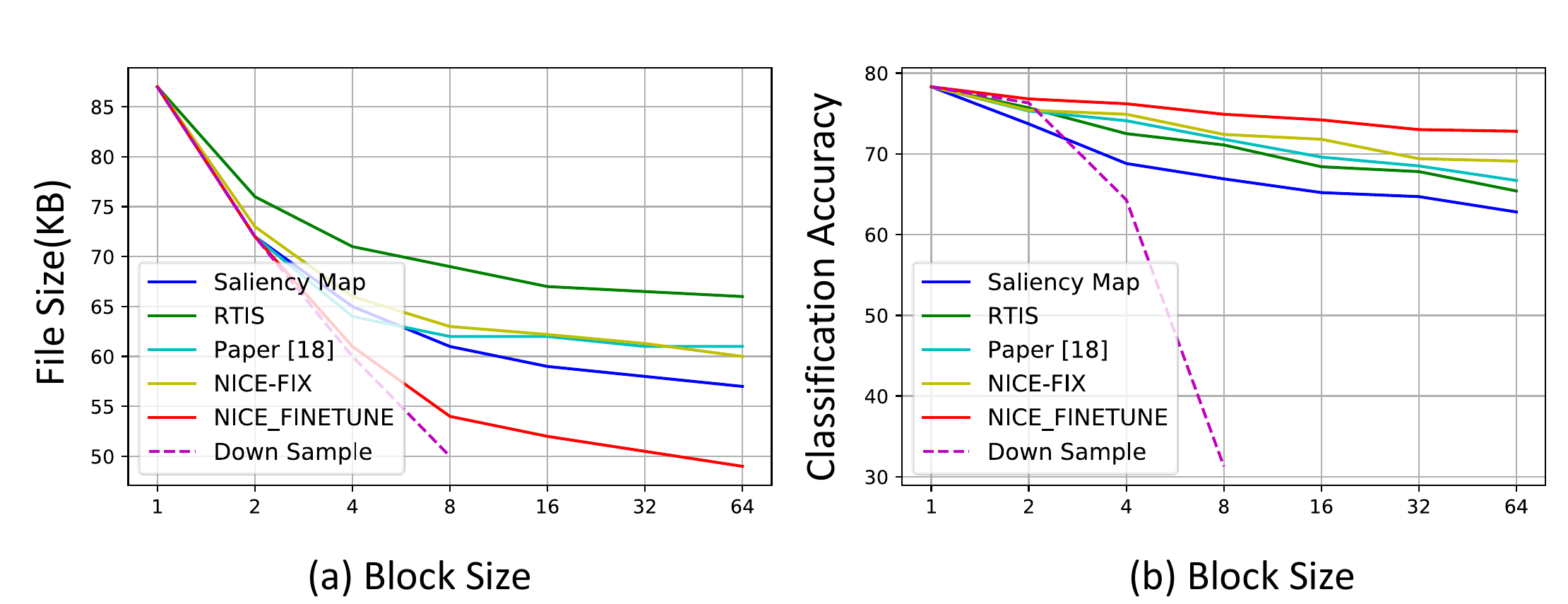}
	\end{center}\vspace{-5pt}
	\caption{The evolution of (a) average file size of the PNG compressed image and (b) classification accuracy as a function of block size $b$ of NICE-fixed, NICE-finetuned, Saliency Map, RTIS, CAM (paper~\cite{Prakash_2017}) and down sampling.}
	\label{fig:curve}\vspace{-5pt}
\end{figure}

To quantitatively evaluate the trade-off between semantic compression rate and classification accuracy,  we train the NICE pipeline with \textbf{Discriminator-fixed} and \textbf{Discriminator-finetuned}~\footnote{For discriminator-finetuned, we set the parameters of Conv-4, Conv-5 and the FC layers of ResNet18 to be trainable and freeze all the other layers.} with $b=16$ to generate sparse masks for each Caltech256 test image. After training, we generate mixed-resolution images with a different $b$ in $\{1,2,4,8,16,32,64\}$. We then use PNG~\cite{sayood2002lossless}, a standard image compression algorithm, to store the generated mixed-resolution images and report the file sizes~\footnote{The reason that we choose PNG~\cite{sayood2002lossless} instead of JPEG~\cite{wallace1992jpeg} for compression is because PNG is a lossless compression. Thus, the file size reduction of the mixed-resolution images can be 100\% attributed to NICE, and the possible artifacts introduced by JPEG, a lossy compression, can be avoided.}. We also classify the mixed-resolution images with the discriminators to report classification accuracies. As a comparison, the same procedure is applied to Saliency Map, RTIS and CAM (paper~\cite{Prakash_2017}) for semantic image compression. We also include a baseline that uses down sampling in our compression pipeline to demonstrate the importance of salient region detection for semantic compression. Specifically, we use down sampling to generate low resolution images regardless of salient regions of the images. When testing the accuracy on the down sampled images, we upsample them to original resolution by bilinear interpolation.

Fig.~\ref{fig:curve} shows the average file size of compressed images and the corresponding classification accuracy as a function of block size $b$. As we can see, when the block size increases, the file size of the compressed images decreases (higher compression rate) and the classification accuracy also decreases (lower classification accuracy), and vis-versa. The classification accuracies of NICE-finetuned are significantly higher than the other four baseline methods, meanwhile it achieves the best compression rate.
When the block-size is 8, NICE-finetuned
achieves a 1.6x compression rate (87KB vs. 54KB) with a small
(3.35\%) accuracy drop (78.30\% vs. 74.95\%), demonstrating the
superior performance of NICE on semantic image compression.

Comparing NICE-finetuned with down sampling, the results show that down sampling hurts classification accuracy significantly because it uniformly drops pixels regardless of their saliency. For example, when we down sample images with a factor of 8, it reduces the average file size significantly (slightly better than NICE), but the corresponding classification accuracy is only 30.83\%, while NICE still achieves an accuracy of 74.95\%. Therefore, the mixed-resolution images produced by NICE can achieve a much better balance between compression rates and final classification accuracies than down sampling. 

Note that semantic image compression rate depends on the size of salient regions of an image. Given the large objects in Caltech256, the 1.6x compression rate means NICE uses 60\% pixels to achieve a similar classification accuracy. To achieve even higher compression rates, other image compression algorithms~\cite{rippel2017real,NakMaeMiy18} can be used to compress NICE generated images further since our algorithm is complementary to these compression techniques.

\subsection{Inference Time Comparison}

\begin{table}[t]
	\begin{tabular}{|c|c|c|c|c|}
		\hline
		\begin{tabular}[c]{@{}c@{}}Explanation \\ Algorithm\end{tabular}    & Saliency Map~\cite{simonyan2013deep} & CAM~\cite{CAM16}  & RTIS~\cite{dabkowski2017real} & NICE \\ \hline
		\begin{tabular}[c]{@{}c@{}}Run Time (sec)\\on 1000 Images\\mean(std)\end{tabular}  & 13.94(0.55)   & 9.87(0.48)   & 1.73(0.02) & 0.59(0.01) \\ \hline
		\end{tabular}
	\caption{Inference time comparison between NICE and the baseline algorithms. The results are averaged over 100 runs.}
	\label{tab:time}\vspace{-10pt}
\end{table}

Besides the improved explanation performance of NICE, another advantage of NICE is its superior inference speed over the competing methods. As discussed above, to generate the sparse mask to explain the decision of a target CNN, NICE only needs one forward propagation of the generator network, while backpropagation-based algorithms, like Saliency Map and CAM, requires heavy computation of backpropagation to generate the salient regions. Similar to NICE, RTIS does not need backpropagation at inference time, but it requires to compute the feature maps of intermediate layers of the target CNN as the input of the generator, which is also time consuming. To have a quantitative speed comparison, we calculate the inference times of different explanation algorithms on 1000 images from Caltech256. We run each experiment 100 times on a NVIDIA Tesla V100 GPU and report the average run-times in Table~\ref{tab:time}. As we can see, NICE is about 23x times faster than Saliency Map and 16.5x faster than CAM, while being 2.8x faster than non-backpropagation based RTIS.

\section{Conclusion}
We propose NICE, a unified end-to-end trainable framework, for neural explanation and semantic image compression. Compared to many existing explanation algorithms that heavily rely on backpropagation, the sparse masks generated by NICE are much more concise and coherent and align well with human intuitions. With the sparse masks, the proposed mixed-resolution image compression further achieves higher compression rates compared to the existing semantic compression algorithms, while retaining similar classification accuracies with the original images. We conduct a series of experiments on multiple image classification benchmarks with multiple CNN architectures and demonstrate its improved explanation quality and semantic image compression rate.

As for future work, we plan to extend the technique developed here to other domains, such as text and bioinformatics for neural explanation and summarization, where interpretable decisions are also critical for the deployment of DNNs.

\section{Acknowledgment}
The authors would like to thank the anonymous reviewers for their comments and suggestions, which helped improve the quality of this paper. The authors would also gratefully acknowledge the support of VMware Inc. for its university research fund to this research.

\bibliographystyle{ieeetr}
\bibliography{egbib}


\begin{IEEEbiography} [{\includegraphics[width=1in]{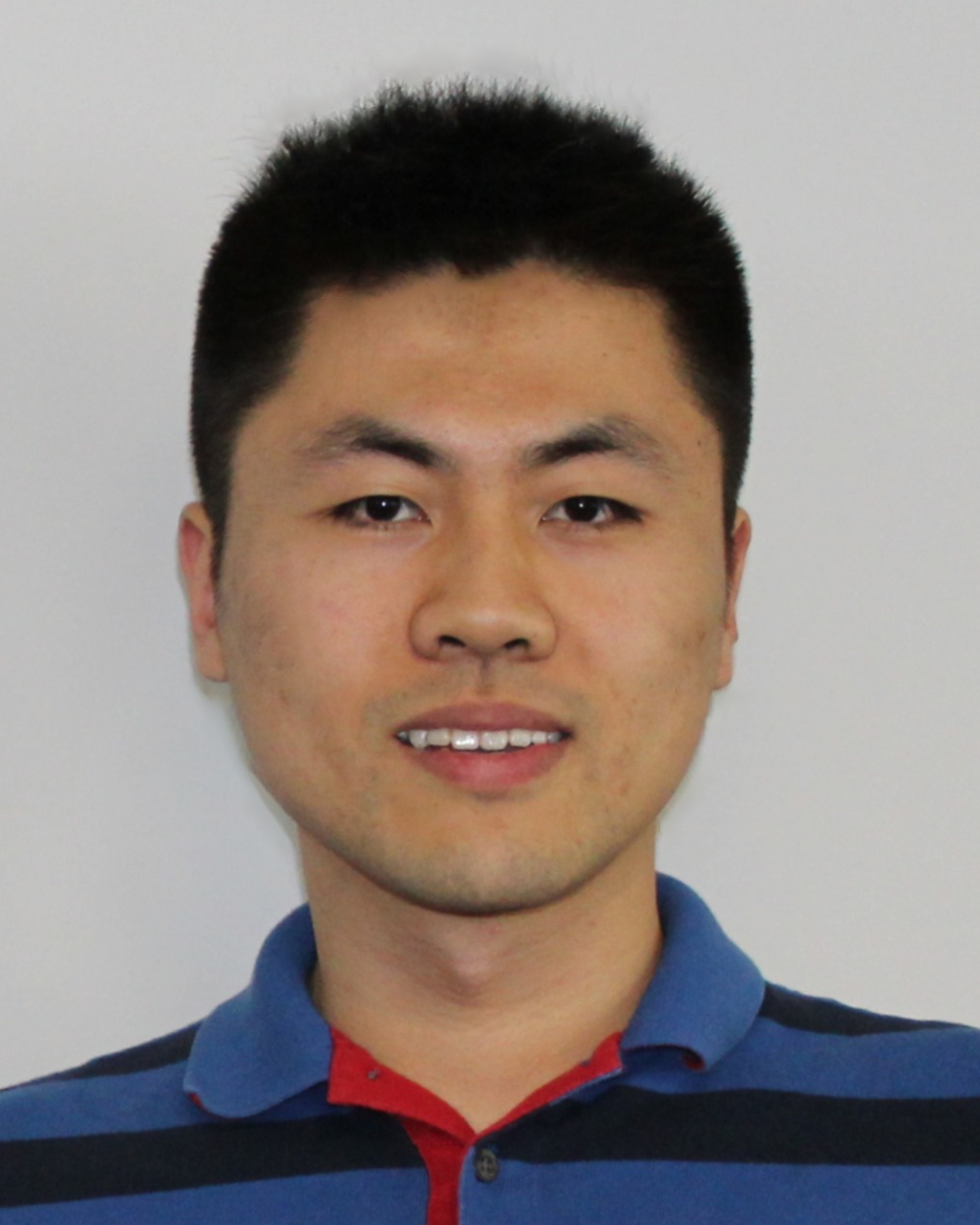}}] {Xiang Li} received the undergraduate
degree in electrical engineering from Donghua University, China, in 2013, and the
master degree in pattern recognition from Northeastern University, China, in 2018. Currently, he is pursuing his PhD degree in computer science at Georgia State University. His research interests 
lie in the robustness of deep learning, security and interpretability of deep neural networks.
\end{IEEEbiography}

\begin{IEEEbiography}[{\includegraphics[width=1in]{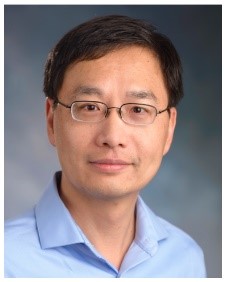}}] {Shihao Ji} (Member, IEEE) is an associate professor in the Computer Science department at Georgia State University. His principal research interests lie in the area of machine learning and deep learning with an emphasis on high-performance computing. He is interested in developing efficient algorithms that can learn from a variety of data sources (e.g., image, audio, and text) on a large scale and automate decision-making processes in dynamic environments. Dr. Ji received his PhD in Electrical and Computer Engineering from Duke University in 2006. After that he was an research associate at Duke for about 1.5 years. Prior to joining GSU, he spent about 10 years in industry research labs.
\end{IEEEbiography}

\EOD

\end{document}